\documentclass[a4paper,11pt]{article}
\usepackage[T1]{fontenc} 
\usepackage{float} 

\usepackage{appendix}

\makeatletter
\gdef\@fpheader{ }
\gdef\@journal{ }
\makeatother
\RequirePackage{amsmath}
\RequirePackage{amssymb}
\RequirePackage{epsfig}
\RequirePackage{graphicx}
\RequirePackage[numbers,sort&compress]{natbib}
\RequirePackage{color}
\RequirePackage[colorlinks=true
,urlcolor=blue
,anchorcolor=blue
,citecolor=blue
,filecolor=blue
,linkcolor=blue
,menucolor=blue
,pagecolor=blue
,linktocpage=true
,pdfproducer=medialab
,pdfa=true
]{hyperref}

\newif\ifnotoc\notocfalse
\newif\ifemailadd\emailaddfalse
\newif\iftoccontinuous\toccontinuousfalse
\makeatletter
\def\@subheader{\@empty}
\def\@keywords{\@empty}
\def\@abstract{\@empty}
\def\@xtum{\@empty}
\def\@dedicated{\@empty}
\def\@arxivnumber{\@empty}
\def\@collaboration{\@empty}
\def\@collaborationImg{\@empty}
\def\@proceeding{\@empty}
\def\@preprint{\@empty}

\newcommand{\subheader}[1]{\gdef\@subheader{#1}}
\newcommand{\keywords}[1]{\if!\@keywords!\gdef\@keywords{#1}\else%
\PackageWarningNoLine{\jname}{Keywords already defined.\MessageBreak Ignoring last definition.}\fi}
\renewcommand{\abstract}[1]{\gdef\@abstract{#1}}
\newcommand{\dedicated}[1]{\gdef\@dedicated{#1}}
\newcommand{\arxivnumber}[1]{\gdef\@arxivnumber{#1}}
\newcommand{\proceeding}[1]{\gdef\@proceeding{#1}}
\newcommand{\xtumfont}[1]{\textsc{#1}}
\newcommand{\correctionref}[3]{\gdef\@xtum{\xtumfont{#1} \href{#2}{#3}}}
\newcommand\jname{JHEP}

\newcommand\preprint[1]{\gdef\@preprint{\hfill #1}}

\makeatother

\newcommand\note[2][]{%
\if!#1!%
\stepcounter{footnote}\footnotetext{#2}%
\else%
{\renewcommand\thefootnote{#1}%
\footnotetext{#2}}%
\fi}

\makeatletter
\newtoks\auth@toks
\renewcommand{\author}[2][]{%
  \if!#1!%
    \auth@toks=\expandafter{\the\auth@toks#2\ }%
  \else
    \auth@toks=\expandafter{\the\auth@toks#2$^{#1}$\ }%
  \fi
}
\makeatother
\makeatletter
\newtoks\affil@toks\newif\ifaffil\affilfalse
\newcommand{\affiliation}[2][]{%
\affiltrue
  \if!#1!%
    \affil@toks=\expandafter{\the\affil@toks{\item[]#2}}%
  \else
    \affil@toks=\expandafter{\the\affil@toks{\item[$^{#1}$]#2}}%
  \fi
}
\makeatother
\makeatletter
\newtoks\email@toks\newcounter{email@counter}%
\newcommand{\emailAdd}[1]{%
\emailaddtrue%
\ifnum\theemail@counter>0\email@toks=\expandafter{\the\email@toks, \@email{#1}}%
\else\email@toks=\expandafter{\the\email@toks\@email{#1}}%
\fi\stepcounter{email@counter}}
\newcommand{\@email}[1]{\href{mailto:#1}{\tt #1}}
\makeatother

\makeatletter
\newcommand*\collaboration[1]{\gdef\@collaboration{#1}}
\newcommand*\collaborationImg[2][]{\gdef\@collaborationImg{#2}}
\makeatletter
\newcommand\afterLogoSpace{\smallskip}
\newcommand\afterSubheaderSpace{\vskip3pt plus 2pt minus 1pt}
\newcommand\afterProceedingsSpace{\vskip21pt plus0.4fil minus15pt}
\newcommand\afterTitleSpace{\vskip23pt plus0.06fil minus13pt}
\newcommand\afterRuleSpace{\vskip23pt plus0.06fil minus13pt}
\newcommand\afterCollaborationSpace{\vskip3pt plus 2pt minus 1pt}
\newcommand\afterCollaborationImgSpace{\vskip3pt plus 2pt minus 1pt}
\newcommand\afterAuthorSpace{\vskip5pt plus4pt minus4pt}
\newcommand\afterAffiliationSpace{\vskip3pt plus3pt}
\newcommand\afterEmailSpace{\vskip16pt plus9pt minus10pt\filbreak}
\newcommand\afterXtumSpace{\par\bigskip}
\newcommand\afterAbstractSpace{\vskip16pt plus9pt minus13pt}
\newcommand\afterKeywordsSpace{\vskip16pt plus9pt minus13pt}
\newcommand\afterArxivSpace{\vskip3pt plus0.01fil minus10pt}
\newcommand\afterDedicatedSpace{\vskip0pt plus0.01fil}
\newcommand\afterTocSpace{\bigskip\medskip}
\newcommand\afterTocRuleSpace{\bigskip\bigskip}
\newlength{\affiliationsSep}
\newcommand\beforetochook{\pagestyle{myplain}\pagenumbering{roman}}

\DeclareFixedFont\trfont{OT1}{phv}{b}{sc}{11}

\renewcommand\maketitle{
\pagestyle{empty}
\thispagestyle{titlepage}
\setcounter{page}{0}
\noindent{\small\scshape\@fpheader}\@preprint\par

\afterLogoSpace
\if!\@subheader!\else\noindent{\trfont{\@subheader}}\fi
\afterSubheaderSpace
\if!\@proceeding!\else\noindent{\sc\@proceeding}\fi
\afterProceedingsSpace
{\LARGE\flushleft\sffamily\bfseries\@title\par}
\afterTitleSpace
\hrule height 1.5\p@%
\afterRuleSpace
\if!\@collaboration!\else
{\Large\bfseries\sffamily\raggedright\@collaboration}\par
\afterCollaborationSpace
\fi
\if!\@collaborationImg!\else
{\normalsize\bfseries\sffamily\raggedright\@collaborationImg}\par
\afterCollaborationImgSpace
\fi
{\bfseries\raggedright\sffamily\the\auth@toks\par}
\afterAuthorSpace
\ifaffil\begin{list}{}{%
\setlength{\leftmargin}{0.28cm}%
\setlength{\labelsep}{0pt}%
\setlength{\itemsep}{\affiliationsSep}%
\setlength{\topsep}{-\parskip}}
\itshape\small%
\the\affil@toks
\end{list}\fi
\afterAffiliationSpace
\ifemailadd 
\noindent\hspace{0.28cm}\begin{minipage}[l]{.9\textwidth}
\begin{flushleft}
\textit{E-mail:} \the\email@toks
\end{flushleft}
\end{minipage}
\else 
\PackageWarningNoLine{\jname}{E-mails are missing.\MessageBreak Plese use \protect\emailAdd\space macro to provide e-mails.}
\fi
\afterEmailSpace
\if!\@xtum!\else\noindent{\@xtum}\afterXtumSpace\fi
\if!\@abstract!\else\noindent{\renewcommand\baselinestretch{.9}\textsc{Abstract:}}\ \@abstract\afterAbstractSpace\fi
\if!\@keywords!\else\noindent{\textsc{Keywords:}} \@keywords\afterKeywordsSpace\fi
\if!\@arxivnumber!\else\noindent{\textsc{ArXiv ePrint:}} \href{http://arxiv.org/abs/\@arxivnumber}{\@arxivnumber}\afterArxivSpace\fi
\if!\@dedicated!\else\vbox{\small\it\raggedleft\@dedicated}\afterDedicatedSpace\fi
\ifnotoc\else
\iftoccontinuous\else\newpage\fi
\beforetochook\hrule
\tableofcontents
\afterTocSpace
\hrule
\afterTocRuleSpace
\fi
\setcounter{footnote}{0}
\pagestyle{myplain}\pagenumbering{arabic}
} 

\renewcommand{\baselinestretch}{1.1}\normalsize
\divide\@tempdima\baselineskip
\@tempcnta=\@tempdima
\skip\@mpfootins = \skip\footins
\renewcommand{\@dotsep}{10000}

\newcommand\ps@myplain{
\pagenumbering{arabic}
\renewcommand\@oddfoot{\hfill-- \thepage\ --\hfill}
\renewcommand\@oddhead{}}
\let\ps@plain=\ps@myplain

\newcommand\ps@titlepage{\renewcommand\@oddfoot{}\renewcommand\@oddhead{}}

\numberwithin{equation}{section}

\renewcommand\section{\@startsection{section}{1}{\z@}%
                                   {-3.5ex \@plus -1.3ex \@minus -.7ex}%
                                   {2.3ex \@plus.4ex \@minus .4ex}%
                                   {\normalfont\large\bfseries}}
\renewcommand\subsection{\@startsection{subsection}{2}{\z@}%
                                   {-2.3ex\@plus -1ex \@minus -.5ex}%
                                   {1.2ex \@plus .3ex \@minus .3ex}%
                                   {\normalfont\normalsize\bfseries}}
\renewcommand\subsubsection{\@startsection{subsubsection}{3}{\z@}%
                                   {-2.3ex\@plus -1ex \@minus -.5ex}%
                                   {1ex \@plus .2ex \@minus .2ex}%
                                   {\normalfont\normalsize\bfseries}}
\renewcommand\paragraph{\@startsection{paragraph}{4}{\z@}%
                                   {1.75ex \@plus1ex \@minus.2ex}%
                                   {-1em}%
                                   {\normalfont\normalsize\bfseries}}
\renewcommand\subparagraph{\@startsection{subparagraph}{5}{\parindent}%
                                   {1.75ex \@plus1ex \@minus .2ex}%
                                   {-1em}%
                                   {\normalfont\normalsize\bfseries}}

\def\fnum@figure{\textbf{\figurename\nobreakspace\thefigure}}
\def\fnum@table{\textbf{\tablename\nobreakspace\thetable}}

\long\def\@makecaption#1#2{%
  \vskip\abovecaptionskip
  \sbox\@tempboxa{\small #1. #2}%
  \ifdim \wd\@tempboxa >\hsize
    \small #1. #2\par
  \else
    \global \@minipagefalse
    \hb@xt@\hsize{\hfil\box\@tempboxa\hfil}%
  \fi
  \vskip\belowcaptionskip}

\renewenvironment{thebibliography}[1]{%
\begin{oldthebibliography}{#1}%
\small%
\raggedright%
\setlength{\itemsep}{5pt plus 0.2ex minus 0.05ex}%
}%
{%
\end{oldthebibliography}%
}

\begin{document}


\title{\boldmath Networks with pixel embedding:
a method to improve noise resistance in image classification}

\author[a]{Yang Liu,}
\author[b,1]{Hai-Long Tu,}\note{Hai-Long Tu and Yang Liu contributed equally to this work.}
\author[a,2]{Chi-Chun Zhou,}\note{Corresponding author. zhouchichun@dali.edu.cn}
\author[a]{Yi Liu}
\author[b]{and Fu-Lin Zhang}


\affiliation[a]{School of Engineering, Dali University, Dali, Yunnan 671003, PR China}
\affiliation[b]{RoyalFlush Information Network Co.,Ltd,  HangZhou ZheJiang 310023, PR China}










\abstract{In the task of image classification, 
usually, the network is sensitive to noises. 
For example, an image of cat with noises might be misclassified as an ostrich.
Conventionally, to overcome the problem of noises, 
one uses the technique of data augmentation, that is, 
to teach the network to distinguish noises by 
adding more images with noises in the training dataset.
In this work, we provide a noise-resistance network in images 
classification by introducing a technique of pixel embedding.
We test the network with pixel embedding,
which is abbreviated as the network with PE,
on the mnist database of handwritten digits.
It shows that the network with PE outperforms the conventional network on  
images with noises. 
The technique of pixel embedding can be used in many tasks of image classification 
to improve noise resistance.}


\maketitle
\flushbottom


\section{Introduction}
In the task of image classification,
information of labels is embedded in pixels of the image. 
For example, in an image of dogs,
a group of pixels with different values will give the typical pattern of the dog, 
such as the ear and the nose.
If the network can remember those typical patterns constructed by a group of pixels, 
then it can predict the label of an image with high accuracy. 
For example, neural networks based on convolution and pooling
are very good at extracting typical patterns from pixels of the image
\cite{krizhevsky2012imagenet,gu2018recent} and even outperform humans
in some tasks of image classification \cite{alom2018history}.

However, one finds that the conventional network is sensitive to noises 
\cite{geirhos2017comparing,da2016empirical}. 
For example, an image of cat with noises might be misclassified as an ostrich.
The low noise resistance of the conventional network will lead to 
problems such as the safety problem of face recognition.
Therefore, how to increase the noise resistance of the network in image classification
becomes a problem.

Conventionally, to overcome the problem of noises, 
one uses the technique of data augmentation \cite{frieden1975image,russo2002image,ray2012noise,ye2004image}, 
that is,
to teach the network to distinguish noises by 
adding more image samples with noises in the training dataset.
Or, one develops techniques of images noise reduction
\cite{vincent2010stacked,zhang2017beyond,lefkimmiatis2017non,brooks2019unprocessing}.

In the task of image classification, instead of understanding the meaning
of pixels, the network just need to
remember the typical patterns constructed by a group of pixels. 
On the one hand, noises are random values added to 
pixels of the image. On the other hand, the conventional network 
is designed to remember typical patterns constructed by a group of pixels.
Thus, noises will greatly interfere in  
those typical patterns that has been learned by the 
network. As a result, networks will give wrong labels because
pixels with noises now become highly confusing. 

In this paper, we provide a noise-resistance network in image 
classification by introducing a technique of pixel embedding.
The network with pixel embedding 
abbreviated as the network with PE is designed not only to remember
the typical pattern
but also to "understand" the meaning of the pixels.
The method in the present paper is enlightened by the 
classical technique of words embedding \cite{levy2014neural,tang2014learning,goldberg2014word2vec} 
in the natural language processing (NLP). 
The technique of words embedding helps the network to "understand"
the meaning of the words and learns knowledges such as analogy and context.
We test the network with PE on the mnist database of handwritten digits.
By comparing the behavior of
networks with and without PE on the test dataset of noised images of the mnist dataset, 
we show that the network with PE outperforms the conventional network on  
images with noises and thus is with high noise resistance.
The technique of pixel embedding can be used in many tasks of image classification 
to improve noise resistance.
The source code is released  
in \href{https://github.com/zhouchichun/pixels_embedding}{github}. 
\note{$https://github.com/zhouchichun/pixels\_embedding$}

The work is organized as follows: in Sec. 2, we introduce the technique of
pixel embedding. In Sec. 3, 
we test the network with PE in the classification task on
the mnist database of handwritten digits. 
Conclusions and outlooks are given in Sec. 4.

\section{pixel embedding}

In this section, we introduce the technique of pixel embedding.
The main ideal of the technique of pixel embedding is to replace
the single-valued pixel with a vector which has the shape $1\times M$ with $M$ the
embedding size. Values of the vector are parameters that need to be trained.
By using the technique of pixel embedding, the network is able to "understand"
the meaning of the pixels and learn to distinguish noises automatically. 

\begin{figure}
\centering
\includegraphics[width=0.9\textwidth]{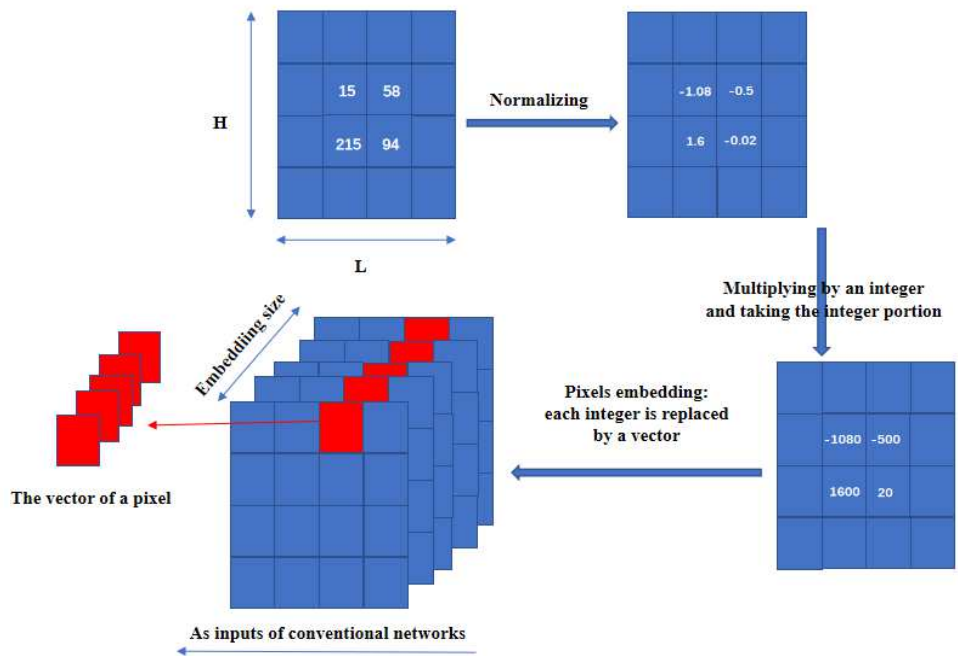}
\caption{The procedure of pixel embedding.}
\label{pe}
\end{figure}

The main procedure of pixel embedding is as follow:
for an image with $H\times L$ pixels, 
firstly, we normalize the pixels by subtracting the average
and dividing the standard deviation.
Secondly, we convert the normalized pixels into an integer by
multiplying pixels by an integer, say $1000$, 
and taking the integer portion.
Thirdly, we introduce an embedding dictionary 
which maps an integer to a unique random vector
with embedding size, say $64$. The procedure is shown in Fig. (\ref{pe}).

\begin{figure}
\centering
\includegraphics[width=0.9\textwidth]{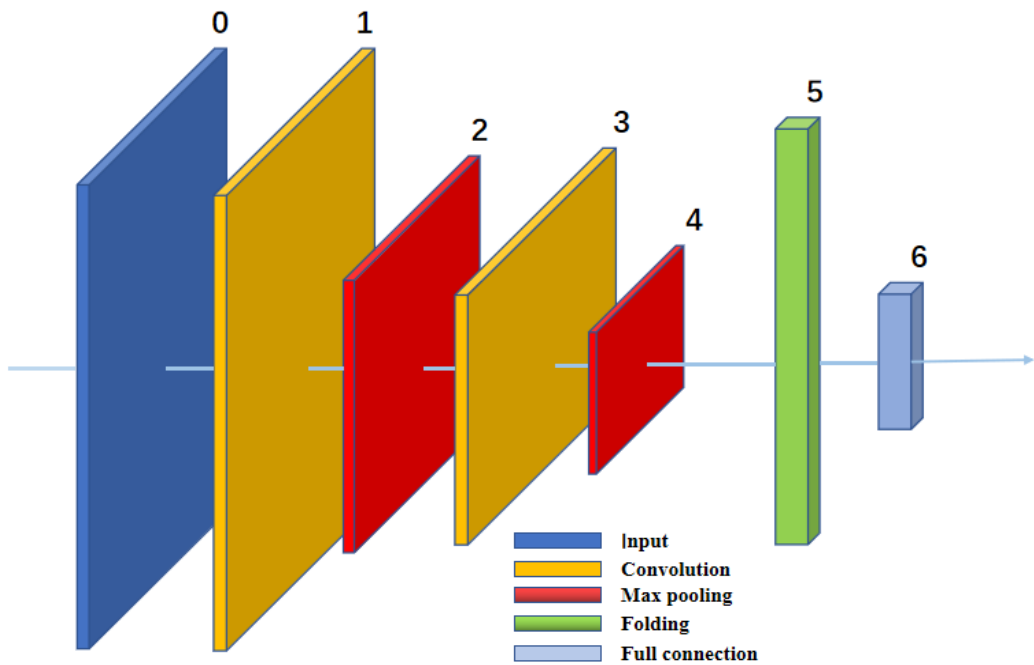}
\caption{The structure of the network used in the present paper. $E\_s$ is short for embedding size.}
\label{vgg16}
\end{figure}

\textit{The network used in the experiment.} 
The technique of pixel embedding can be used in various kinds of conventional networks.
Without loss of generality, the network we used here is based on the VGG16, 
a classical network for image classification. The structure of the network is shown in Fig. (\ref{vgg16}).
The detail of the structure of the neural network is given in Table. (\ref{VGG16_t})

\begin{table}[H]
\label{VGG16_t}
\caption{ConvNet configurations.}
\centering
\begin{tabular}{|c||c||c||c||c|}  
\hline   
$\text{Layer}$ & $\text{Name}$ & $\text{Output Shape}$\\  
\hline   
$0$ & $\text{Input}$ & $\text{(28,28,$E_{s}$)}$\\
\hline   
$1$& $\text{Convolution}$& $\text{(28,28,64)}$\\
\hline   
$2$& $\text{Max pooling}$& $\text{(14,14,64)}$\\
\hline   
$3$& $\text{Convolution}$& $\text{(14,14,64)}$\\
\hline   
$4$& $\text{Max pooling}$& $\text{(7,7,64)}$\\
\hline   
$5$& $\text{Folding}$& $\text{(1,3136)}$\\
\hline   
$6$& $\text{Full connection}$& $\text{(1,10)}$\\
\hline 
\end{tabular}
\label{VGG16_t}
\end{table}

Training parameters are show in Table. (\ref{train})
\begin{table}[H]
\label{train}
\caption{Training parameters of the network used in the present paper.}
\centering
 \begin{tabular}{|c||c|} 
\hline   
$\text{Length of the embedding dictionary }$&$2000$\\ 
\hline 
$\text{Optimization algorithm  }$& $\text{Adam }$\\ 
\hline 
$\text{Integer to multiply}$&$1000$\\ 
\hline   
$\text{Embedding size }$&$64$\\
\hline   
$\text{Learning rate }$& $0.0001$ \\
\hline     
$\text{Batch size}$& $64$ \\
\hline 
\end{tabular}
\label{train}
\end{table}

\section{The classification task on the mnist database}
In this section, we test the network with PE on the mnist database.
The mnist database of handwritten digits \cite{lecun1998mnist,deng2012mnist} is an famous open dataset for image
classification task. 
We train networks with PE and without PE on $60,000$ training images respectively 
and test them on $10,000$ test images with different noises.
The behaviors of networks with and without PE on the test dataset of noised images of the mnist dataset are as follows.

\textit{For the test dataset with gauss noise}, the result is given in Table. (\ref{table_gauss}).
\begin{table}[H]
\caption{The accuracy on test dataset with gauss noise of networks with and without PE.}
\centering
 \begin{tabular}{|c||c||c||c|}  
\hline   
$\text{Number}$&$\text{Index}$&$\text{Accuracy (without PE)}$&$\text{Accuracy (with PE)}$\\  
\hline   
$\text{case1}$&$\text{Standard test dataset}$& $98.7$& $97.83$ \\  
\hline 
$\text{case2}$&$\mu=1.0,\sigma=1.0$& $85.70$& $95.94$\\  
\hline   
$\text{case3}$&$\mu=5.0,\sigma=1.0$& $88.66$& $88.10$\\ 
\hline
$\text{case4}$&$\mu=10,\sigma=2$& $78.53$& $80.18$\\  
\hline  
$\text{case5}$&$\mu=25,\sigma=5$& $53.24$& $67.92$\\ 
\hline 
$\text{case6}$&$\mu=50,\sigma=10$& $28.71$& $59.89$\\
\hline 
$\text{case7}$&$\mu=75,\sigma=15$& $18.61$& $40.22$\\ 
\hline 
$\text{case8}$&$\mu=100,\sigma=10$& $17.81$& $33.52$\\
\hline 
$\text{case9}$&$\mu=150,\sigma=10$& $12.22$& $18.84$\\
\hline
\end{tabular}
\label{table_gauss}
\end{table}

\begin{figure}[H]
\centering
\includegraphics[width=0.9\textwidth]{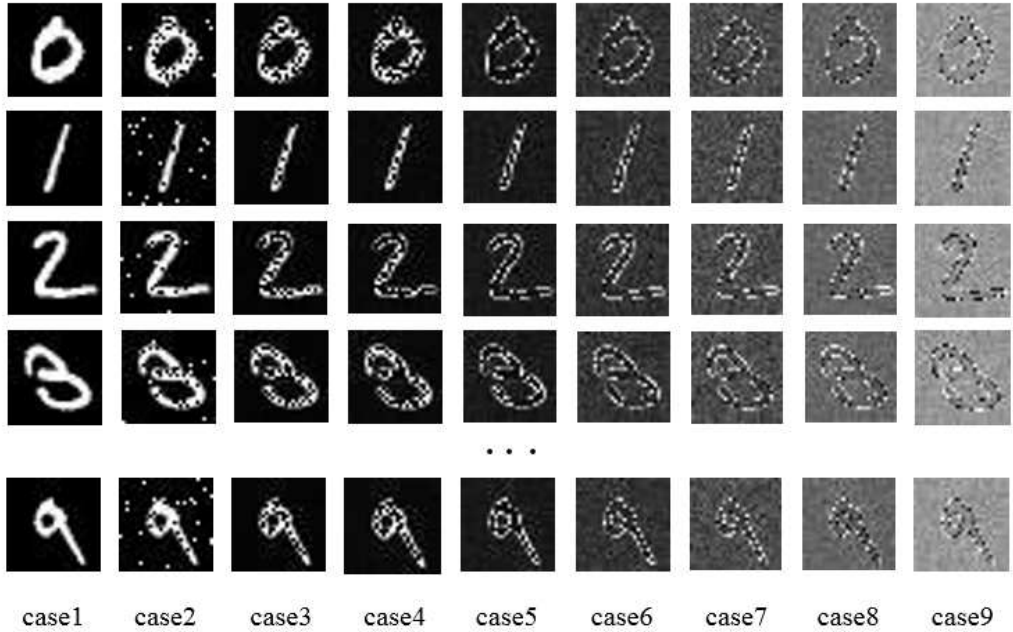}
\caption{The accuracy on test dataset with gauss noise of networks with and without PE.}
\label{gauss}
\end{figure}

In Table. (\ref{table_gauss}), "$\mu=1.0,\sigma=1.0$" means that the average value of noises is $1.0$
and the standard deviation is $1.0$ and so on. Because the pixels are integers 
between $0$ and $255$, the random noise are transformed to integers 
between $0$ and $255$. As shown in Fig. (\ref{gauss}).

\begin{figure}
\centering
\includegraphics[width=0.9\textwidth]{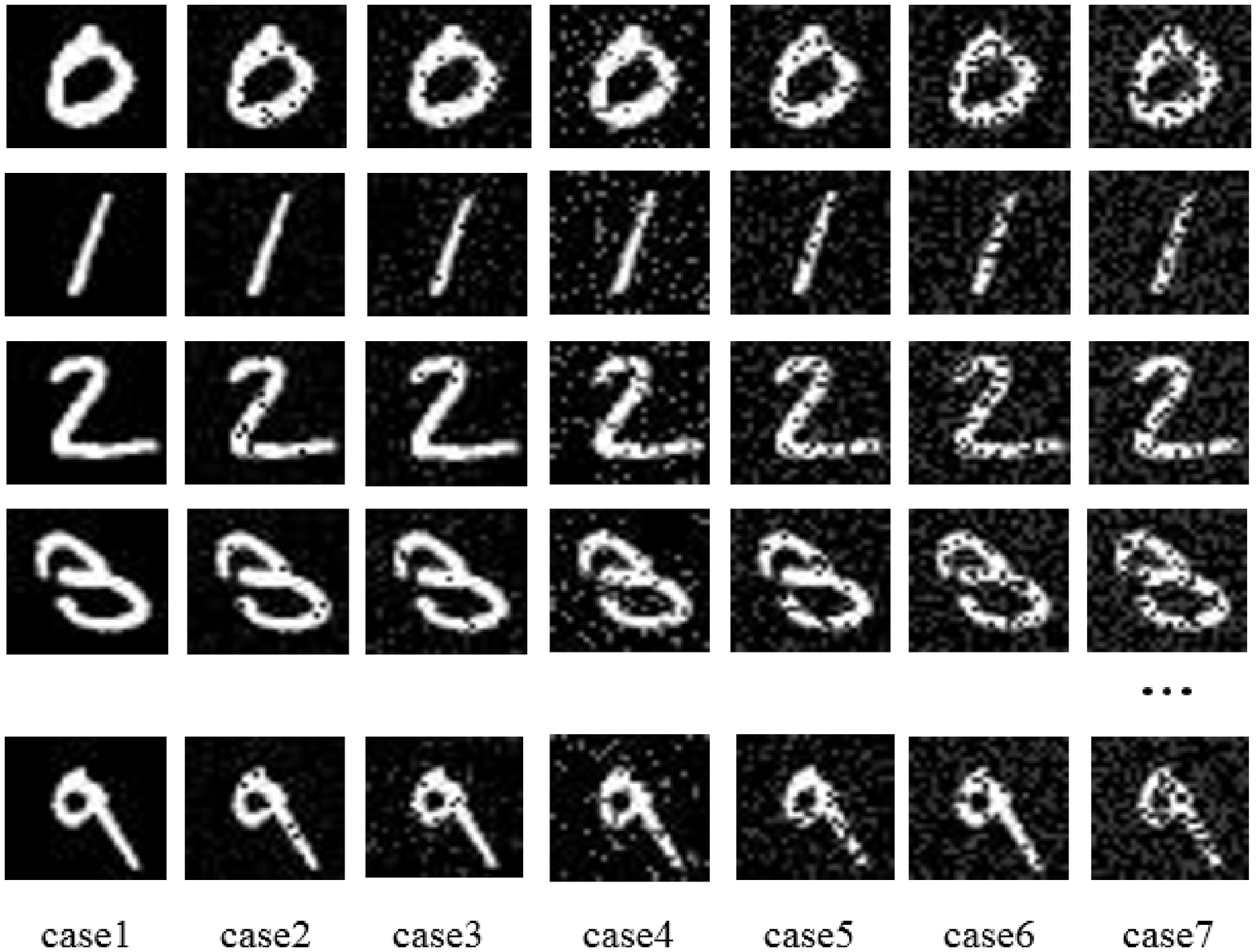}
\caption{The accuracy on test dataset with salt-and-pepper noise of networks with and without PE.}
\label{jiaoyan1}
\end{figure}

\begin{figure}
\centering
\includegraphics[width=0.9\textwidth]{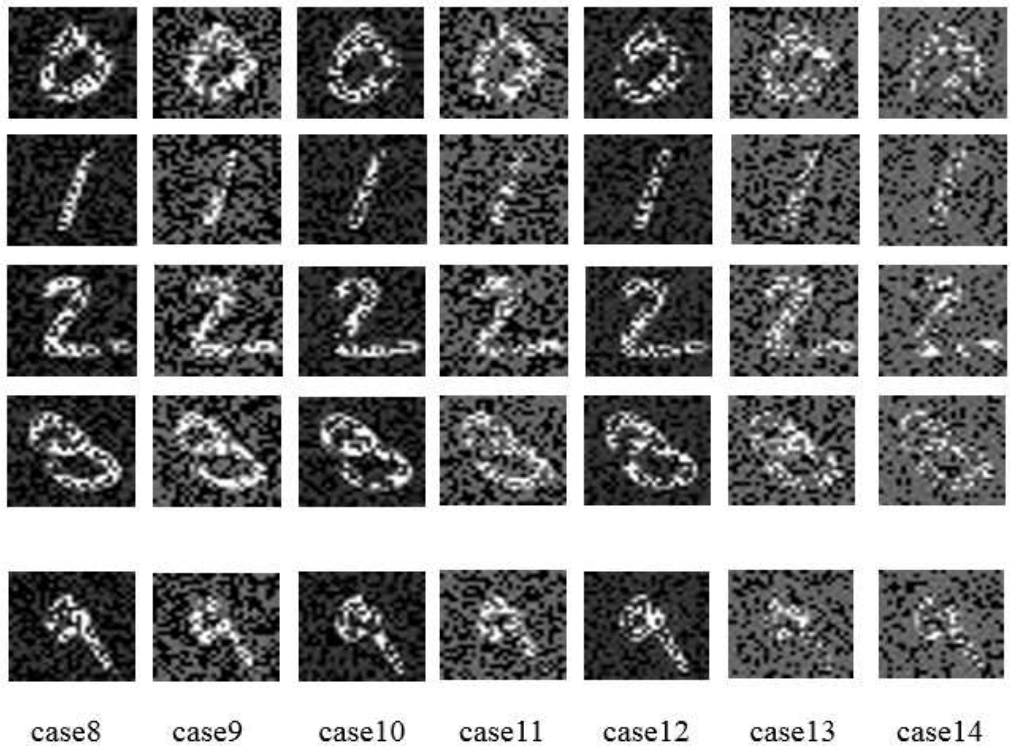}
\caption{The accuracy on test dataset with salt-and-pepper noise of networks with and without PE.}
\label{jiaoyan2}
\end{figure}

\textit{For the test dataset with salt-and-pepper noise}, the result is given in Table. (\ref{table_jiaoyan}).
\begin{table}[H]
\caption{The accuracy on test dataset with salt-and-pepper noise of networks with and without PE.}
\centering
 \begin{tabular}{|c||c||c||c||c|}   
\hline   
$\text{Number}$&$\text{Index}$&$\text{Accuracy (without PE)}$&$\text{Accuracy (with PE)}$\\ 
\hline
$\text{case1}$&$\text{Standard test dataset }$& $98.17$& $97.83$\\
\hline
$\text{case2}$&$p=0.1,v=25$& $94.72$& $96.96$\\
\hline
$\text{case3}$&$p=0.1,v=50$& $90.10$& $97.07$\\
\hline
$\text{case4}$&$p=0.1,v=100$& $84.24$& $96.03$\\
\hline
$\text{case5}$&$p=0.2,v=50$& $86.93$& $95.28$\\ 
\hline
$\text{case6}$&$p=0.3,v=50$& $82.36$& $91.14$\\ 
\hline
$\text{case7}$&$p=0.4,v=50$& $77.31$& $89.59$\\
\hline
$\text{case8}$&$p=0.5,v=50$& $69.53$& $78.51$\\ 
\hline
$\text{case9}$&$p=0.5,v=100$& $49.47$& $77.75$\\
\hline
$\text{case10}$&$p=0.6,v=50$& $60.00$& $81.55$\\
\hline
$\text{case11}$&$p=0.6,v=100$& $35.38$& $64.12$\\
\hline
$\text{case12}$&$p=0.7,v=50$& $48.40$& $71.72$\\
\hline
$\text{case13}$&$p=0.7,v=100$& $22.20$& $59.09$\\
\hline
$\text{case14}$&$p=0.8,v=100$& $14.93$& $47.62$\\
\hline
\end{tabular}
\label{table_jiaoyan}
\end{table}
In Table. (\ref{table_jiaoyan}), "$p=0.1,v=25$" means that the value of the noise pixels is $25$
and the probability of a pixel to have a noise is $0.1$ and so on. 
The random noise are also transformed to integers 
between $0$ and $255$. As shown in Figs. (\ref{jiaoyan1})-(\ref{jiaoyan2}).

\section{Conclusions and outlooks}

In this paper, beyond the conventional method of data augmentation,
where more sample datas are added to the training dataset, 
we propose a new method that lead to a noise-resistance network where
no more training sample is needed. 
In this approach, a technique of
pixel embedding is introduced. This method is enlightened by the 
classical techniques of word embedding in the NLP. 
The technique of pixel embedding 
extends the meaning of pixels, and thus, the network can learn the knowledge
such as the noise, the context, and the meaning of the pixel.
We test the network with PE on the mnist database of handwritten digits.
By comparing the behavior of
networks with and without PE on the test dataset of noised images of the mnist dataset, 
we show that the network with PE outperforms the conventional network on  
images with noises and thus is with high noise resistance.
The technique of pixel embedding
can be used in various kinds of tasks of image classification 
to improve noise resistance.

\textit{To remember or to learn.} In the task of image classification,
instead of understanding the meaning
of pixels, the network just need to
remember the typical patterns constructed by a group of pixels. 
To remember is much easier than to "understand", thus, 
the methodology of deep neural network achieve success
in the classification tasks of images first. 
However, at cases, such as image classification with noises,
understanding a sentence in the NLP,
discovering hidden laws from datas in financial, mathematics, and physics,
the network has to learn more knowledges from the 
raw data other than just remember typical patterns constructed by
combinations of raw features. 
To do this, 
we have to introduce techniques that can make the network more "intelligent".

The technique of pixel embedding is an attempt to make the network
to "understand" the meaning of pixels other than to remember typical patterns 
constructed by groups of pixels.

\section{Authors' contributions}
Hai-Long Tu performed the experiment.
Yang Liu checked the algorithm, reconstructed the code, and 
repeated the experiment.
Hai-Long Tu and Yang Liu contributed equally to this work

\section{Acknowledgments}
We are very indebted to Profs. Wu-Sheng Dai, Guan-Wen Fang, and Yong Xie for their encouragements. 
This work is supported by "Artificial intelligence + 
domestic waste classification" 
of the basic research youth project 
of Yunnan Provincial Science and Technology Department.

\section{Appendix}
It shows that the pixel embedding increases the noise resisting ability of the algorithm.
However, in the experiment, the pixel embedding is applied after normalization preprocess.
Here, in order to show that the pixel embedding contributes mainly to the noise resisting ability
other than the normalization preprocess,
we give the result of network without pixel embedding but with normalization, as shown in Table. (\ref{table_gauss1}) 

\begin{table}[H]
\caption{The accuracy on test dataset with gauss noise of networks 
with only normalization preprocess.}
\centering
 \begin{tabular}{|c||c||c|}  
\hline   
$\text{Number}$&$\text{Index}$&$\text{Accuracy}$\\  
\hline   
$\text{case1}$&$\text{Standard test dataset}$& $97.57$ \\  
\hline 
$\text{case2}$&$\mu=1.0,\sigma=1.0$& $70.25$\\  
\hline   
$\text{case3}$&$\mu=5.0,\sigma=1.0$& $72.26$\\ 
\hline
$\text{case4}$&$\mu=10,\sigma=2$& $54.14$\\  
\hline  
$\text{case5}$&$\mu=25,\sigma=5$& $24.85$\\ 
\hline 
$\text{case6}$&$\mu=50,\sigma=10$& $11.56$\\
\hline 
$\text{case7}$&$\mu=75,\sigma=15$& $10.11$\\ 
\hline 
$\text{case8}$&$\mu=100,\sigma=10$& $9.94$\\
\hline 
$\text{case9}$&$\mu=150,\sigma=10$& $9.93$\\
\hline
\end{tabular}
\label{table_gauss1}
\end{table}












\end{document}